\documentclass[runningheads]{llncs}

\usepackage[mobile]{eccv}

\usepackage{eccvabbrv}

\usepackage{graphicx}
\usepackage{booktabs}
\usepackage{multirow} % for \multirow in tables
\usepackage{caption}  % for \captionof{table} inside minipage

\usepackage[accsupp]{axessibility}  % Improves PDF readability for those with disabilities.

\usepackage{wrapfig}
\usepackage{float}
\usepackage{hyperref}

\usepackage{orcidlink}
\definecolor{alizarin}{rgb}{0.82, 0.1, 0.26}
\hypersetup{
    colorlinks=true,  
    citecolor=black  
}

\begin{document}

\newcommand{\xiaodong}[1]{{\textcolor{orange}{[xd: #1]}}}
\newcommand{\xd}[1]{{\textit{\textcolor{orange}{{#1}}}}}

% Add a period to the end of an abbreviation unless there's one
% already, then \xspace.
\makeatletter
\DeclareRobustCommand\onedot{\futurelet\@let@token\@onedot}
\def\@onedot{\ifx\@let@token.\else.\null\fi\xspace}

\def\eg{\emph{e.g}\onedot} \def\Eg{\emph{E.g}\onedot}
\def\ie{\emph{i.e}\onedot} \def\Ie{\emph{I.e}\onedot}
\def\cf{\emph{c.f}\onedot} \def\Cf{\emph{C.f}\onedot}
\def\etc{\emph{etc}\onedot} \def\vs{\emph{vs}\onedot}
\def\wrt{w.r.t\onedot} \def\dof{d.o.f\onedot}
\def\etal{\emph{et al}\onedot}
\makeatother

\title{CutClaw: Agentic Hours-Long Video Editing via Music Synchronization} 

\titlerunning{CutClaw: Agentic Hours-Long Video Editing via Music Synchronization}

\author{Shifang Zhao\inst{1,2}\and
Yihan Hu\inst{2}\and
Ying Shan\inst{3}\and
Yunchao Wei\inst{1\dag}\and
Xiaodong Cun\inst{2\dag}}

\begingroup
\renewcommand{\thefootnote}{}
\footnotetext{$^\dag$ Corresponding authors.}
\endgroup

\authorrunning{Zhao et al.}

\institute{Beijing Jiaotong University \and
GVC Lab, Great Bay University \and
ARC Lab, Tencent  \\
{\hypersetup{urlcolor=alizarin}\url{https://github.com/GVCLab/CutClaw}}
}

\maketitle
\vspace{-2.5em}
\begin{figure*}[h]
\centering
  \includegraphics[width=1\textwidth]{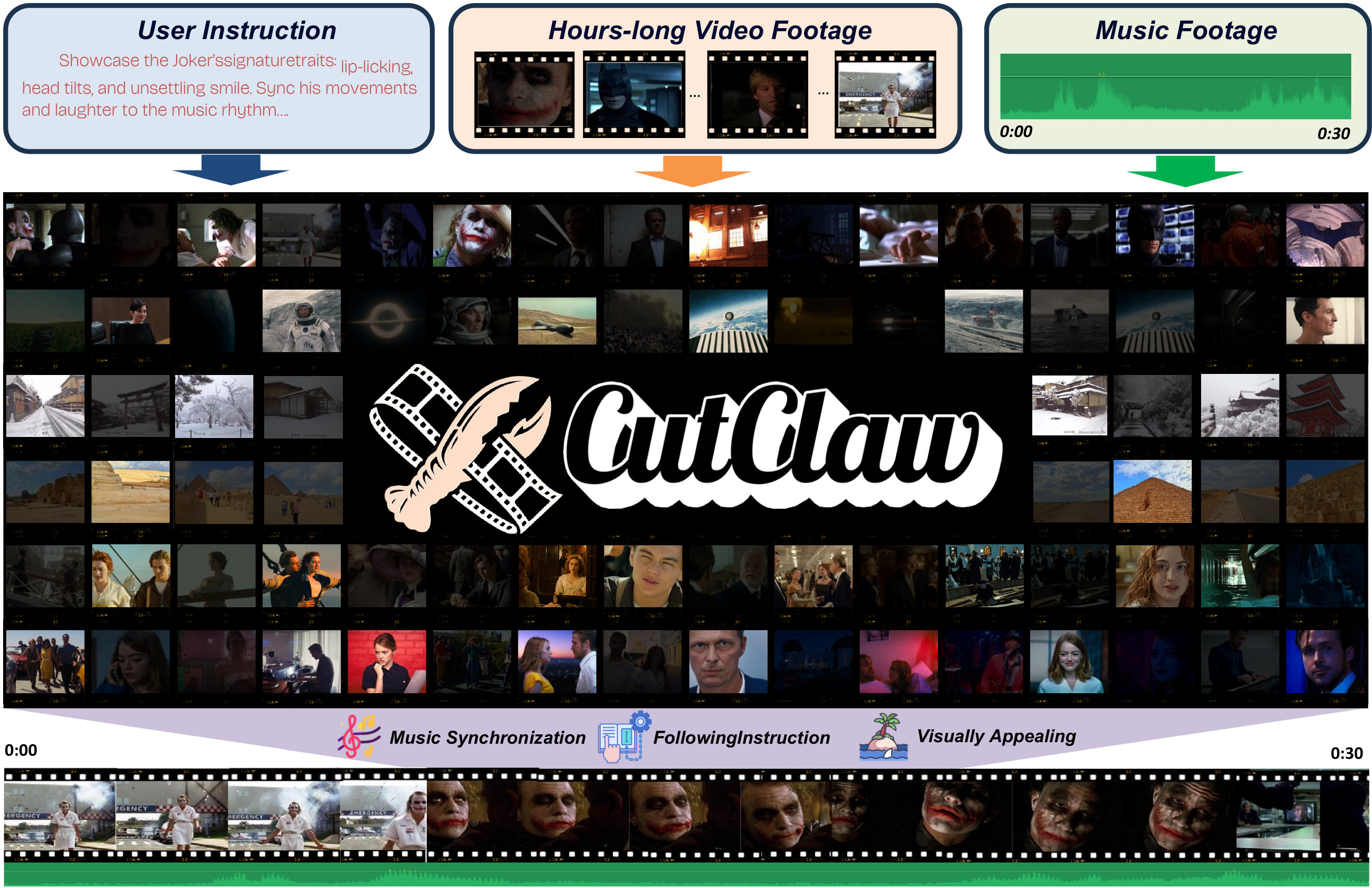}
  \vspace{-2em}
    \caption{We present an automated music-driven video editing system that transforms hours-long footage into high-quality short videos based on user instructions and given music rhythm, so that the resulting video demonstrates precise music synchronization, faithful instruction following, and visually appealing aesthetics.}
  \label{fig:teaser}
\end{figure*}
\vspace{-3.5em}
\begin{abstract}
Editing the video content with audio alignment forms a digital human-made art in current social media.
However, the time-consuming and repetitive nature of manual video editing has long been a challenge for filmmakers and professional content creators alike. 
In this paper, we introduce CutClaw, an autonomous multi-agent framework designed to edit hours-long raw footage into meaningful short videos that leverages the capabilities of multiple Multimodal Language Models~(MLLMs) as an agent system. It produces videos with synchronized music, followed by instructions, and a visually appealing appearance. 
In detail, our approach begins by employing a hierarchical multimodal decomposition that captures both fine-grained details and global structures across visual and audio footage.
Then, to ensure narrative consistency, a Playwriter Agent orchestrates the whole storytelling flow and structures the long-term narrative, anchoring visual scenes to musical shifts. 
Finally, to construct a short edited video, Editor and Reviewer Agents collaboratively optimize the final cut via selecting fine-grained visual content based on rigorous aesthetic and semantic criteria. 
We conduct detailed experiments to demonstrate that CutClaw significantly outperforms state-of-the-art baselines in generating high-quality, rhythm-aligned videos. The code is available at: {\hypersetup{urlcolor=alizarin}\url{https://github.com/GVCLab/CutClaw}}.
\end{abstract}
%---------------------------------------------
\section{Introduction}
\label{sec:intro}

% Video editing\footnote{In this article, video editing means that we trim the raw footage for better storytelling; the terms ``cutting'' and ``editing'' are used interchangeably.} is arguably the most transformative stage of storytelling, where the ``invisible art'' turns chaotic reality into coherent narrative. While the ubiquity of high-definition cameras has made capturing hours of raw footage effortless, the cognitive load of post-production has grown exponentially. Professional editing is not merely about trimming timeline sequences; it is a complex optimization process that requires maintaining a global narrative arc while strictly adhering to local rhythmic constraints, \eg, matching visual energy with auditory beats. For a human editor, distilling hours of raw material into a compelling minute video requires sifting through massive unstructured data to find ``needle-in-a-haystack'' moments, a task that is notoriously labor-intensive and demands high-level aesthetic intuition.
Videos are inherently multimodal, weaving together visual and auditory streams. Consequently, 
Audio-driven video editing\footnote{In this work, ``editing'' and ``cutting'' are used interchangeably to denote the temporal selection and assembly of raw video segments.} represents the most transformative stage of storytelling, fusing sight and sound into organic harmony.
Moving beyond simple temporal concatenation, cinematic editing is inherently a complex multimodal alignment problem. In practice, distilling hours of untrimmed video into a concise output requires traversing a massive search space to retrieve sparse, salient segments that simultaneously advance the global storyline and strictly adhere to local auditory dynamics. Balancing the dual constraints of maintaining global narrative coherence and ensuring fine-grained visual-audio harmony renders professional editing a highly labor-intensive process that is heavily dependent on human aesthetic intuition.

% Existing approaches to automated video editing generally fall into three categories: template-based methods, highlight detection methods, and text-based editing methods. Template-based methods~\cite{pr,dafineqi, capcut_official_en} force users into rigid, pre-defined structures with musical beats, resulting in repetitive outputs that lack adaptability. Highlight detection methods~\cite{uvcom} focus on identifying high-quality clips based on visual cues but often treat them in isolation, producing disjointed ``slideshows.'' Finally, text-based approaches~\cite{narratoai} leverage temporally aligned transcripts, prioritizing verbal semantics while often neglecting visual aesthetics. Consequently, current methods struggle to generate cohesive videos with both clear narratives and visual appeal.

Despite recent progress, existing automated video editing frameworks typically neglect the critical role of audio, falling into three suboptimal paradigms. \eg, Template-based methods~\cite{pr,dafineqi, capcut_official_en} force clips into rigid, predefined temporal slots and overlay background music; lacking audio-visual synchronization and semantic awareness, they yield repetitive outputs devoid of narrative progression. Highlight detection methods~\cite{uvcom} optimize for local visual salience but are audio-agnostic, treating clips in isolation and failing to construct a globally coherent narrative. Text-based approaches~\cite{narratoai} prioritize linguistic semantics by aligning visuals with transcripts, yet neglect the underlying musical structure, disrupting both kinetic rhythm and affective energy. Consequently, these methods optimize audio, video, and text instruction independently, struggling to achieve the holistic multimodal alignment required to satisfy the dual constraint of global storytelling and fine-grained visual-audio harmony.

% Despite the surging demand for automation, developing a system that bridges the gap between simple clip assembly and professional-grade storytelling faces three fundamental challenges. \textit{(i) video editing requires understanding super-long contexts}. Professional editing often involves sifting through hours of raw materials to find relevant content, a volume of data that overwhelms the context windows of current multimodal models~\cite{chatgpt,google2025gemini3} and makes it difficult to maintain a global understanding without suffering from information overload. \textit{(ii) existing tools lack narrative awareness}. While traditional methods might successfully detect specific objects like a ``smiling face'', they struggle to arrange shots into a logical sequence that adheres to a user's creative intent, often failing to select clips that are semantically meaningful to the unfolding story~\cite{narratoai}. \textit{(iii) achieving precise visual-audio consistency without templates is difficult}. Professional editing involves ``cutting on the beat'' and matching visual energy with auditory emotion; reproducing this organically requires aligning fine-grained visual content to structural shifts in the audio rather than just mechanical beat matching. 

 To build a system capable of practical audio-visual storytelling entails three fundamental technical challenges. \textit{(i) Context Length Limitation.} The dense visual information required for fine-grained understanding across hours-long raw footage physically surpasses the context window length of current MLLMs~(Multimodal Language Models)~\cite{chatgpt,google2025gemini3}.\textit{(ii) Context-Grounded Storytelling.} Crafting a cohesive visual story requires reconciling external user instructions with the intrinsic semantics of the raw video and audio. It is highly challenging to synthesize a narrative logic that strictly executes creative intent without decoupling from the native context and subjects of the source materials. \textit{(iii) Fine-Grained Cross-Modal Alignment.} Achieving organic visual-audio harmony demands fine-grained temporal grounding to synchronize musical shifts with a holistic understanding of visual plot, aesthetics, and emotion.

To address these challenges, we introduce \textit{CutClaw}, an autonomous MLLM-powered multi-agent framework that mimics a professional post-production workflow through a collaborative, coarse-to-fine hierarchy. To overcome the context length limitation, a \textit{Bottom-Up Multimodal Footage Deconstruction} module abstracts both raw video and audio into structured semantic units of visual scenes and musical sections, enabling both narrative comprehension and fine-grained analysis. To achieve context-grounded storytelling, a \textit{Playwriter} agent acts as a global planner. Using the musical structure as an invariant temporal anchor, it aligns user instructions with the abstracted scenes to synthesize a narrative that executes creative intent while respecting the source material's intrinsic plot. Finally, to achieve fine-grained cross-modal alignment, \textit{Editor} agent and \textit{Reviewer} agent collaboratively perform top-down hierarchical visual grounding. Guided by the summarized script, the \textit{Editor} localizes precise segments, and the \textit{Reviewer} enforces a multi-criteria validity gate to rigorously evaluate plot relevance, visual aesthetics, and instruction following, thereby guaranteeing organic audio-visual harmony.

Our key contributions are summarized as: 
\begin{itemize} 
\vspace{-0.5em}

\item We tackle the novel task of audio-driven video editing, formally modeling it as a joint optimization problem that simultaneously satisfies instruction-driven storytelling and fine-grained rhythmic harmony.

\item We introduce \textbf{CutClaw}, an MLLM-powered multi-agent framework that tackles the computationally intractable search space of hours-long footage. It integrates \textit{bottom-up multimodal deconstruction} with a collaborative agentic workflow, where a \textit{Playwriter} orchestrates music-anchored narrative planning, while \textit{Editor} and \textit{Reviewer} agents collaboratively execute precise segment selection.

\item Extensive experiments and user studies demonstrate that CutClaw significantly outperforms state-of-the-art baselines in visual quality, instruction following, and rhythmic harmony. 
\end{itemize}

%---------------------------------------------

\vspace{-1em}
\section{Related Work}
% This structure is excellent. It moves from the broad task (Editing) to the specific technical enabler (Grounding) to the modern solution architecture (Agents).

% Here is the drafted content for the Related Work section, tailored to support your introduction's logic.

\noindent\textbf{AI-assisted Video Editing.} Video editing has evolved from optimization-based heuristics to data-driven frameworks. Early pioneering works, such as Write-A-Video~\cite{write_a_video} and ESA~\cite{esa_editing}, formulated editing as an energy minimization problem to align shots with themed cues. Recent generative methods ~\cite{generative_timelines, text_to_edit_ad} have shifted towards constructing visual sequences driven by high-level instructions or subtitle narratives~\cite{narratoai}. However, these methods are fundamentally limited to assembling pre-segmented clips, rely on explicit scripts for narrative structure, and critically neglect the rhythmic guidance of the music modality. In contrast, CutClaw directly processes raw, untrimmed footage without manual scripts, formulating editing as a hierarchical narrative construction that simultaneously guarantees semantic storytelling and fine-grained audio-visual harmony.

\noindent\textbf{Video Temporal Grounding and Highlight Detection.}
Video Temporal Grounding(VTG) and Highlight Detection serve as fundamental prerequisites for editing by determining where to cut within raw footage. VTG aims to localize specific segments based on natural language queries; conventional approaches~\cite{guo2024trace, mu2024snag} rely on pretrained feature encoders, while recent methods~\cite{timer1} leverage MLLMs to enhance instruction understanding. Similarly, Highlight Detection has evolved from using visual saliency scores~\cite{sun2014ranking, xu2021cross, xiong2019less} to incorporating textual prompts~\cite{sun2024tr, uvcom} for better alignment with user preferences. However, both streams of research face significant limitations in professional editing contexts: they struggle to effectively model the long-term context of raw footage and lack precise control over the duration of retrieved results. Consequently, these methods are ill-suited for high-precision audio-visual synchronization tasks, where visual cuts must rigorously align with musical beats and rhythmic patterns. To bridge this gap, we take a step to deal with hours-long video footage with both textual and musical input.

\noindent\textbf{Agents for Video Generation and Editing.} The advent of MLLMs has catalyzed the use of multi-agent collaborations in the video domain~\cite{zhu2025paper2video, liu2024intelligent}. Recent frameworks employ agents for various settings, ranging from generative role-playing in ViMax~\cite{vimax} to non-linear editing in EditDuet~\cite{editduet} and targeted video trimming~\cite{agent_trimming}. However, these systems face critical bottlenecks in scalability and precision. They are constrained by context windows when processing hours-long footage and fail to achieve audio-visual synchronization due to coarse LLM planning. CutClaw overcomes these limitations by pairing a Hierarchical Decomposition strategy for long-context processing with Audio-Anchor Alignment for precise multi-modal synchronization.
\section{Method}
% 1. 最上面放两个音频和视频，视频先放footage再放caption和标号(代表经过deconstruction的结果).
% 2. 写剧本的音频短一些，只保留64附近的就行
% 3. storytelling 上的scene变成带图片的
% 4. 输出最后回到storytelling上
% 5. 每个视频都标注一下时长或者fps

\subsection{Problem Formulation}
Given raw video footage, a target music track, and a text instruction as multimodal inputs, we formulate video editing as an agent-driven segment extraction and assembly problem. By leveraging multiple specialized models and agents, our framework extracts and synchronizes relevant clips to ensure the final output strictly follows the narrative instruction while achieving organic audio-visual harmony.

\begin{figure*}[t]
\centering
  \includegraphics[width=\textwidth]{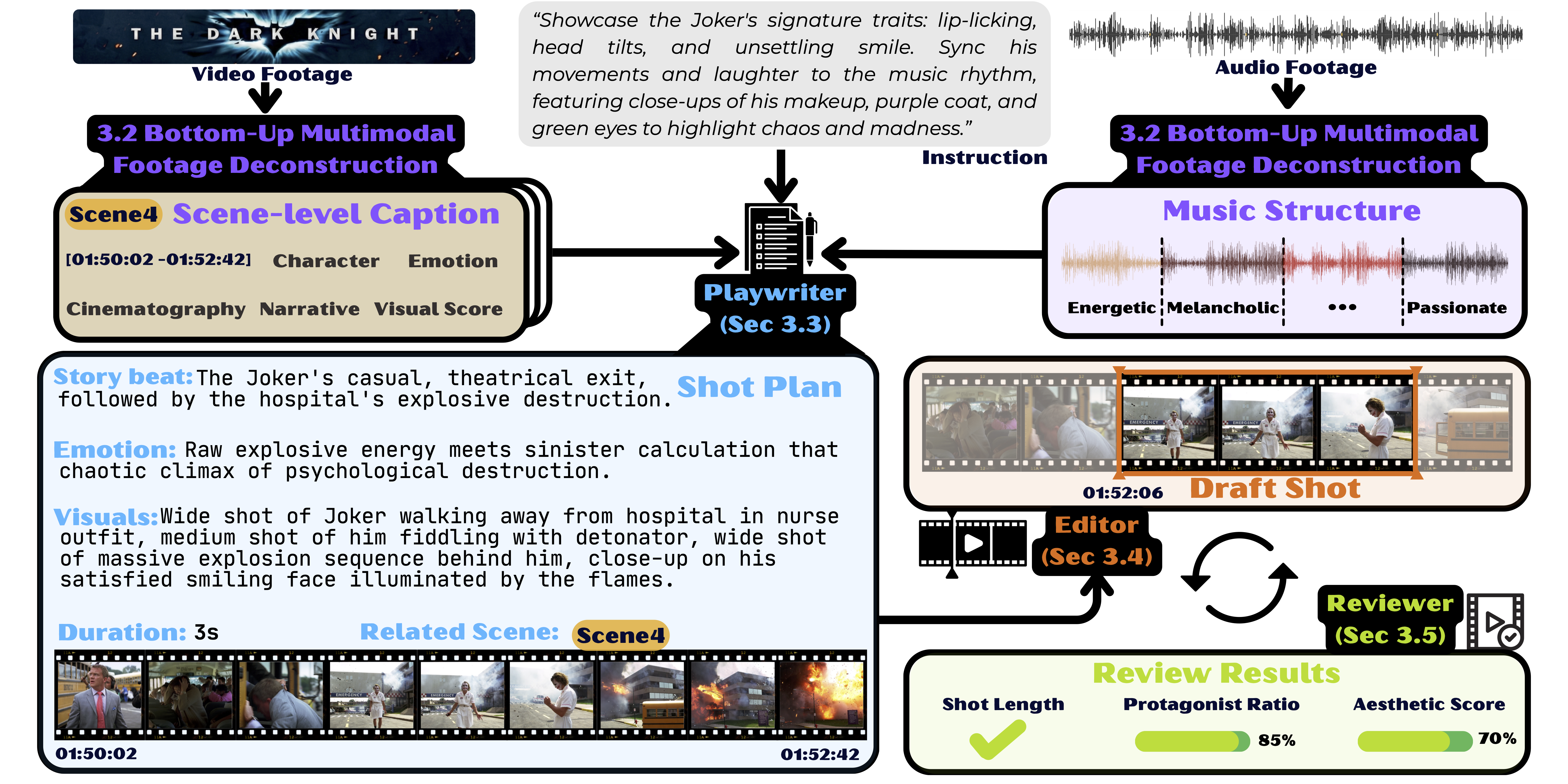}
  \vspace{-2em}
    \caption{\textit{\textbf{The whole workflow of the CutClaw.}} The multi-modal footage is first \textbf{\textcolor[RGB]{120,85,242}{Deconstructed}}, and then, the shot plan is generated by the \textbf{\textcolor[RGB]{96,180,249}{Playwriter}}, scene retrieval and editing by the \textbf{\textcolor[RGB]{219,122,55}{Editor}}, and quality validation by the \textbf{\textcolor[RGB]{187,221,77}{Reviewer}}. }
  \vspace{-2em}
  \label{fig:method}
\end{figure*}

% Central to this process is the progressive alignment of three timelines: the \textbf{Video Footage Timeline} (scene structure parsed via Visual Semantic Aggregation), the \textbf{Storytelling Timeline} (textual narrative generated by the Playwright), and the \textbf{Edited Timeline} (final video assembled from selected footage). The \protect\textcolor[RGB]{41,60,245}{blue arrows} indicate successful execution paths, with adjacent icons denoting tool invocations. Conversely, \protect\textcolor[RGB]{247,206,205}{red arrows} denote failure loops upon rejection. The \protect\textcolor[RGB]{188,221,77}{green lines} trace the mapping from source footage to the final edited timeline. \xiaodong{need update}

Formally, given raw video footage $\mathcal{V}$, the background music track $\mathcal{M}$, and the user instructions $\mathcal{I}$, the target edited video is recomposed by a trimed timeline $\mathcal{E} = (c_1, \dots, c_N)$, which consists of a sequence of clips and each clip $c_i = (t_{i}^{\mathrm{in}}, t_{i}^{\mathrm{out}})$ represents a continuous segment extracted from the original video footage $\mathcal{V}$. We optimize a timeline $\mathcal{E}^*$ to maximize a joint objective function:
    \begin{equation}
    \label{eq:joint_obj}
    \begin{split}
        \mathcal{E}^* = \mathop{\arg\max}_{\mathcal{E}} \Big( 
            & \lambda_v \mathcal{Q}_{\mathrm{vis}}(\mathcal{E}) + 
              \lambda_n \mathcal{Q}_{\mathrm{narr}}(\mathcal{E}) + \\
            & \lambda_c \mathcal{Q}_{\mathrm{cond}}(\mathcal{E}, \mathcal{I}) + \lambda_s \mathcal{Q}_{\mathrm{sync}}(\mathcal{E}, \mathcal{M})
        \Big),
    \end{split}
    \end{equation}
where $\mathcal{Q}_{\mathrm{vis}}$~(\textit{Visual Quality}) ensures aesthetic appeal and protagonist prominence; $\mathcal{Q}_{\mathrm{narr}}$~(\textit{Narrative Flow}) enforces coherent storytelling between adjacent clips; $\mathcal{Q}_{\mathrm{cond}}$~(\textit{Semantic Alignment}) measures the fidelity of selected content to the instructions $\mathcal{I}$; and $\mathcal{Q}_{\mathrm{sync}}$~(\textit{Rhythmic Alignment}) encourages visual cuts to synchronize with musical beats in $\mathcal{M}$. Instead of brute-force searching, we approximate the solution via a \textit{hierarchical search space analysis and pruning strategy}. As shown in Fig.~\ref{fig:method}, we first discretize the high-dimensional footage into structured semantic~(Sec.~\ref{sec:footage parsing}), effectively reducing the solution space. Subsequently, the \textit{Playwriter} (Sec.~\ref{sec:Playwriter}) leverages audio-visual correlations to constrain the search scope to localized candidate pools, enabling the \textit{Editor} (Sec.~\ref{sec:editor}) and \textit{Reviewer} (Sec.~\ref{sec:Reviewer}) to perform efficient fine-grained retrieval and rigorous rejection sampling to finalize the timeline $\mathcal{E}^*$.
    
    % \xd{We analyze the whole system as several parts. including the }
    % To address the computational intractability of optimizing Eq.~\eqref{eq:joint_obj} over the continuous domain of $\mathcal{V}$, we approximate the solution via a hierarchical search space pruning strategy. Instead of brute-force searching, we first discretize the high-dimensional footage into structured semantic units through \textbf{Bottom-Up Multimodal Footage Deconstruction} (Sec.~\ref{sec:footage parsing}), effectively reducing the solution space. Subsequently, the \textbf{Playwriter} (Sec.~\ref{sec:Playwriter}) leverages audio-visual correlations to constrain the search scope to localized candidate pools, enabling the \textbf{Editor} (Sec.~\ref{sec:editor}) and \textbf{Reviewer} (Sec.~\ref{sec:Reviewer}) to perform efficient fine-grained retrieval and rigorous rejection sampling to finalize the timeline $\mathcal{E}^*$.

% \input{vis/fig/video/video}
\begin{figure*}[t]
\centering
  \includegraphics[height=0.5\textwidth]{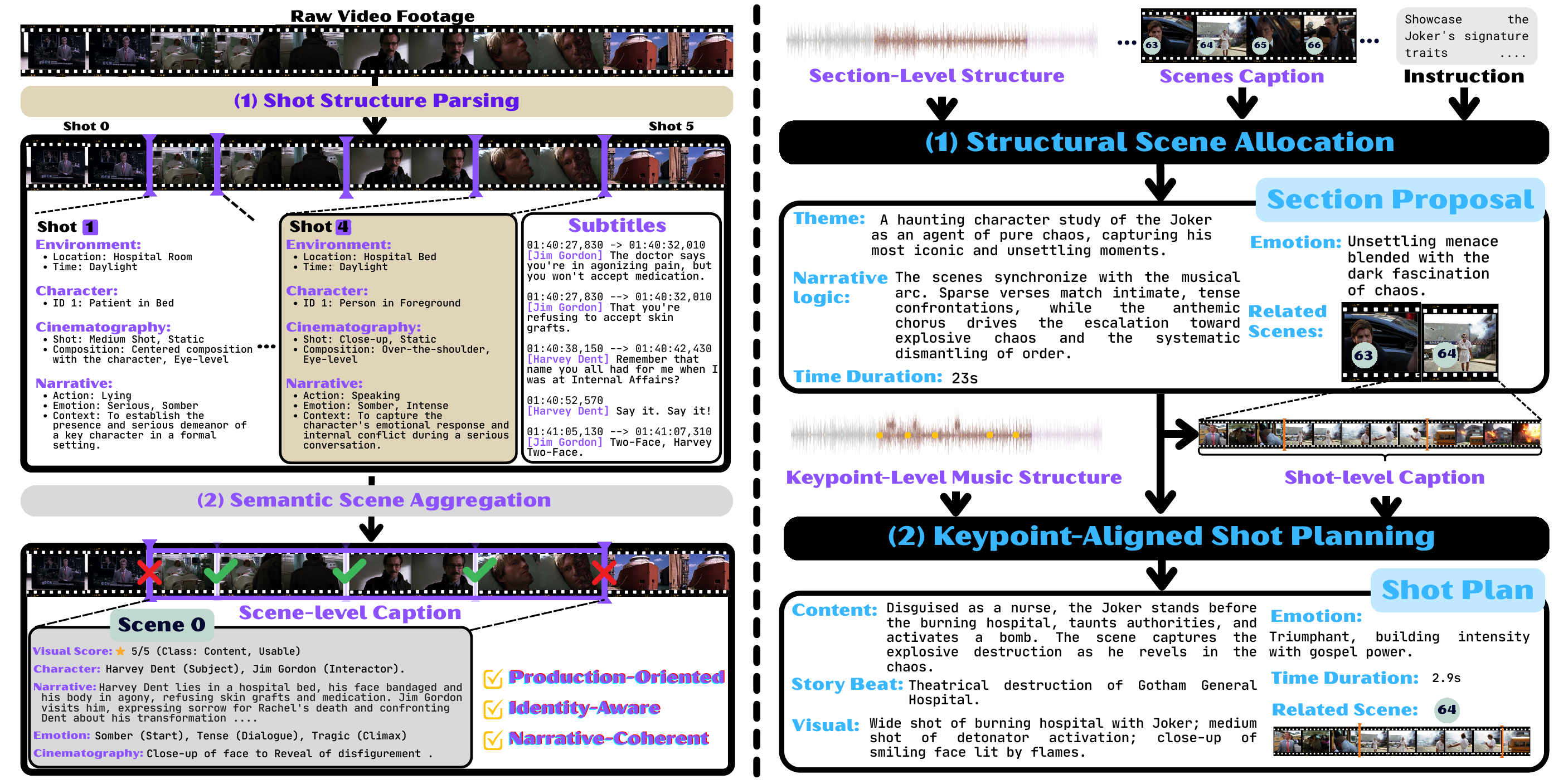}
  \vspace{-2em}
  \caption{\textbf{\textit{Left: Video Shots Aggregation.}} We first perform shot detection on the entire video and conduct a caption for a detailed understanding. Then, we use an LLM to aggregate similar content for a scene-level description. \textbf{\textit{Right: The workflow of Playwriter.}}  Playwriter generate the whole storyline according to the input, and gives the detailed shot plan of each specific shot of music.
  }
  \label{fig:video_playwriter}
  \vspace{-2em}
\end{figure*}

\subsection{Bottom-Up Multimodal Footage Deconstruction}
\label{sec:footage parsing}

The raw footage $\mathcal{V}$ and background music $\mathcal{M}$ are continuous, high-dimensional streams, making direct timeline optimization computationally intractable. To address this, we perform a bottom-up deconstruction to discretize these inputs into structured semantic units, establishing a finite, searchable candidate space for the subsequent hierarchical planning.

\subsubsection{Video Shots Aggregation: From Shot to Scene}
\label{sec:Video parsing} 

Effective editing requires both fine-grained and coarse-grained narrative comprehension. To reconcile these granularity requirements within the context window limits of MLLMs~\cite{Qwen3-VL}, we propose a hierarchical aggregation strategy (Fig.~\ref{fig:video_playwriter} Left). Specifically, we discretize the footage $\mathcal{V}$ into atomic \emph{shots} $\mathcal{S}$ defined as fundamental visual units bounded by camera cuts, which are subsequently aggregated into \emph{scenes} $\mathcal{Z}$ forming contiguous, spatio-temporally coherent shot sequences.

\vspace{-1em}
\paragraph{Shot Parsing and Scene Aggregation.}
To instantiate this hierarchy, we first obtain the atomic shots $\mathcal{S}$ using boundary detection~\cite{pyscenedetect}. For each shot $s_i$, we extract semantic attributes $\mathcal{A}(s_i)$ covering cinematography, character dynamics, and environment via an MLLM~\cite{Qwen3-VL}. To group these individual shots into the defined scenes, we compute a transition similarity $\mathrm{Sim}(s_i, s_{i+1}) = \boldsymbol{\alpha}^{\top}\,\mathbf{v}_{i,i+1}$ between adjacent shots. Here, $\mathbf{v}_{i,i+1}$ denotes the attribute-wise similarity vector derived from the LLM~\cite{reimers-2020-multilingual-sentence-bert} features, and $\boldsymbol{\alpha}$ represents the weight vector balancing the importance of different attributes. A scene boundary is induced whenever this similarity score drops below a predefined threshold $\tau$, effectively partitioning the continuous footage into discrete, meaningful narrative blocks.

\vspace{-1em}
\paragraph{Character-Aware Grounding.}
To ensure narrative consistency involving recurring protagonists, we implement a identity injection. We first analyze the dialogue to infer character identities $\mathcal{H}$ (names and roles). These identities are injected as textual conditioning into the MLLM~\cite{Qwen3-Omni} during scene analysis. This grounds the generated descriptive summary $\mathcal{D}(z_j)$ in specific personas (e.g., replacing ``a man'' with ``Joker''), facilitating reliable cross-scene character tracking.

\subsubsection{Structural Audio Parsing}
\label{sec:Audio parsing}

To maximize Rhythmic Alignment ($\mathcal{Q}_{\mathrm{sync}}$), we convert the continuous music waveform $\mathcal{M}$ into a discrete grid of potential cut points. We employ a hierarchical strategy that bridges micro-level rhythm (beats) with macro-level musical form (sections), providing the Playwriter with rigid temporal anchors.

\vspace{-1em}
\paragraph{Hierarchical Keypoint Detection.}
We first extract perceptually salient \emph{Sound Keypoints} $\mathcal{K}$ on a discrete time axis $\mathcal{T}$~\cite{madmom}. We identify three types of candidates: (i) Downbeats $\mathcal{K}_{\mathrm{db}}$ (bar-level accents); (ii) Pitch Changes $\mathcal{K}_{\mathrm{pc}}$ (melodic transitions); and (iii) Spectral Energy Changes $\mathcal{K}_{\mathrm{se}}$ (timbral transitions). We form a unified candidate pool $\mathcal{K}_{0} = \mathcal{K}_{\mathrm{db}} \cup \mathcal{K}_{\mathrm{pc}} \cup \mathcal{K}_{\mathrm{se}}$ and apply temporal filtering $\Phi(\cdot)$ (e.g., peak de-duplication) to obtain robust boundaries $\mathcal{K} = \Phi(\mathcal{K}_{0})$.

\vspace{-1em}
\paragraph{Structure-Guided Refinement.}
To organize these keypoints, we use an MLLM~\cite{Qwen3-Omni} to partition the track into coarse structural units $\mathcal{U} = \{u_j\}_{j=1}^{M}$ (e.g., verse, chorus). Within each unit $u_j$, we score the contained keypoints $t \in \mathcal{K} \cap u_j$ to retain only the most significant boundaries. The significance score is computed as a weighted sum of cue intensities:
\begin{equation}
    \mathrm{score}(t) = \boldsymbol{\beta}^{\top}\,\mathbf{i}(t), \text{ where } \mathbf{i}(t) = [\mathrm{int}_{\mathrm{db}}(t), \mathrm{int}_{\mathrm{pc}}(t), \mathrm{int}_{\mathrm{se}}(t)]^\top.
\end{equation}
where $\mathrm{int}_{*}(t)$ denotes the intensity of each respective type at time $t$, and $\boldsymbol{\beta}$ is the weight vector. Finally, we generate structure-aligned captions, describing local rhythm, emotion, and energy to guide the visual matching.

\subsection{Playwriter: Music-Anchored Script Synthesis}
\label{sec:Playwriter}

% With the high-dimensional inputs decomposed into semantic scenes $\mathcal{Z}$ and structural audio units $\mathcal{U}$, we transform the complex timeline generation problem from a continuous search into a discrete combinatorial planning task. Thus, 
Given the decomposed semantic scenes $\mathcal{Z}$ and structural audio units $\mathcal{U}$, Playwriter~\cite{google2025gemini3} utilizes the musical structure $\mathcal{U}$ as the invariant temporal anchor for storytelling(Fig.~\ref{fig:video_playwriter} Right). By strictly grounding the visual narrative progression onto this auditory skeleton, the Playwriter enforces Rhythmic Alignment ($\mathcal{Q}_{\mathrm{sync}}$) while optimizing for Instruction Fidelity ($\mathcal{Q}_{\mathrm{cond}}$) and Narrative Flow ($\mathcal{Q}_{\mathrm{narr}}$).
% Instead of exploring the unconstrained video volume, the Playwriter functions as a constraint-aware planner. 
It utilizes structural scene allocation and keypoint-aligned shot planning to map the video scenes $\mathcal{Z}$ onto the musical structure $\mathcal{U}$, generating a shot plan subject to strictly formalized execution rules to guarantee validity:

\begin{enumerate}
\item \textit{Disjoint Resource Allocation (Non-Overlap):} To prevent temporal redundancy, the Playwriter strictly partitions the scene $\mathcal{Z}$. Let $\mathcal{Z}_{u_j} \subset \mathcal{Z}$ denote the subset of scenes allocated to the $j$-th musical unit. For any two distinct units $u_j, u_k \in \mathcal{U}$, we enforce: $ \mathcal{Z}_{u_j} \cap \mathcal{Z}_{u_k} = \emptyset.$ This exclusive assignment ensures that no source material is reused across different narrative blocks, logically satisfying the global non-overlap constraint by construction.

\item \textit{Structural Temporal Anchoring (Music Duration):} To enforce the duration constraint, the generated shot plan $P_j$ for each unit $u_j$ inherits the fixed temporal topology of the audio. The total planned duration is strictly anchored to the audio interval length: $ \sum_{p \in P_j} \text{Duration}(p) \equiv |u_j|.$
\end{enumerate}
Below, we give the details workflow.
\vspace{-1em}
% By internalizing these hard constraints into the generation policy, the Playwriter produces an executable blueprint that is structurally valid, allowing the downstream Editor to focus purely on maximizing visual aesthetic quality ($\mathcal{Q}_{\mathrm{vis}}$). Below, we give the details workflow.

\subsubsection{Structural Scene Allocation}
    The first stage constructs a global mapping between musical structural units and visual scenes. Let $\mathcal{U} = \{u_j\}_{j=1}^{M}$ denote the set of musical units derived in Sec.~\ref{sec:Audio parsing}. The agent generates a structure proposal $\mathcal{P}$ that assigns a subset of candidate scenes $\mathcal{Z}_{u_j} \subset \mathcal{Z}$ to each unit $u_j$.
    
    The allocation is formulated as a conditional generation task:
    \begin{equation}
        \mathcal{Z}_{u_j} = \Phi_{\mathrm{macro}}(u_j, \mathcal{I} \mid \mathcal{Z}),
    \end{equation}
    where $\Phi_{\mathrm{macro}}$ represents the LLM-based~\cite{google2025gemini3} planning function conditioned on the user instruction $\mathcal{I}$. To satisfy the hard temporal constraints, we enforce a strict disjoint set requirement:
    \begin{equation}
        \mathcal{Z}_{u_j} \cap \mathcal{Z}_{u_k} = \emptyset, \quad \forall j \neq k.
    \end{equation}
    If the generated proposal violates this condition (i.e., a scene is reused across different musical sections), the system rejects $\mathcal{P}$ and triggers a regeneration with negative constraints.
\vspace{-1em}
\subsubsection{Keypoint-Aligned Shot Planning}
The second stage refines the allocation into a sequence of executable specifications. For each unit $u_j$, let $\{k_1, \dots, k_L\}$ be the set of fine-grained musical segments contained within its temporal scope. The agent generates a shot plan consisting of specifications $\{p_1, \dots, p_L\}$.

Critically, rather than outputting final timestamps, each specification $p_i = (\tau_i, z_{\mathrm{id}}, d_i)$ serves as a \textit{retrieval constraint} for the subsequent editing phase:
\begin{description}
    \item[$\tau_i$:] The target duration constraint derived directly from the audio segment $k_i$, ensuring rhythmic synchronization ($\mathcal{Q}_{\mathrm{sync}}$).
    \item[$z_{\mathrm{id}}$:] The source scene index selected from $\mathcal{Z}_{u_j}$, which restricts the retrieval search space to the allocated narrative block.
    \item[$d_i$:] A semantic visual description (e.g., specific plot or emotion) that guides the content matching within scene $z_{\mathrm{id}}$.
\end{description}

This hierarchical binding transforms the global optimization problem into a series of local retrieval tasks. By explicitly binding the $i$-th shot to a specific scene $z_{\mathrm{id}}$, the Playwriter effectively prunes the search space for the downstream Editor, ensuring that the final clip selection is conducted successfully.

\subsection{Editor: Top-Down Hierarchical Visual Grounding}
\label{sec:editor}

Operating within the structural shot plan constrained by the Playwriter, the Editor performs fine-grained temporal grounding to determine the precise continuous coordinates of the final timeline $\mathcal{E}^*$. We instantiate the Editor as a ReAct~\cite{yao2022react} agent designed to iteratively maximize the local energy terms of the joint objective function (Eq.~\ref{eq:joint_obj}), specifically targeting Visual Quality ($\mathcal{Q}_{\mathrm{vis}}$).

\begin{wrapfigure}{br}{0.6\columnwidth}
    % \vspace{-15pt} 
    \includegraphics[width=0.6\columnwidth]{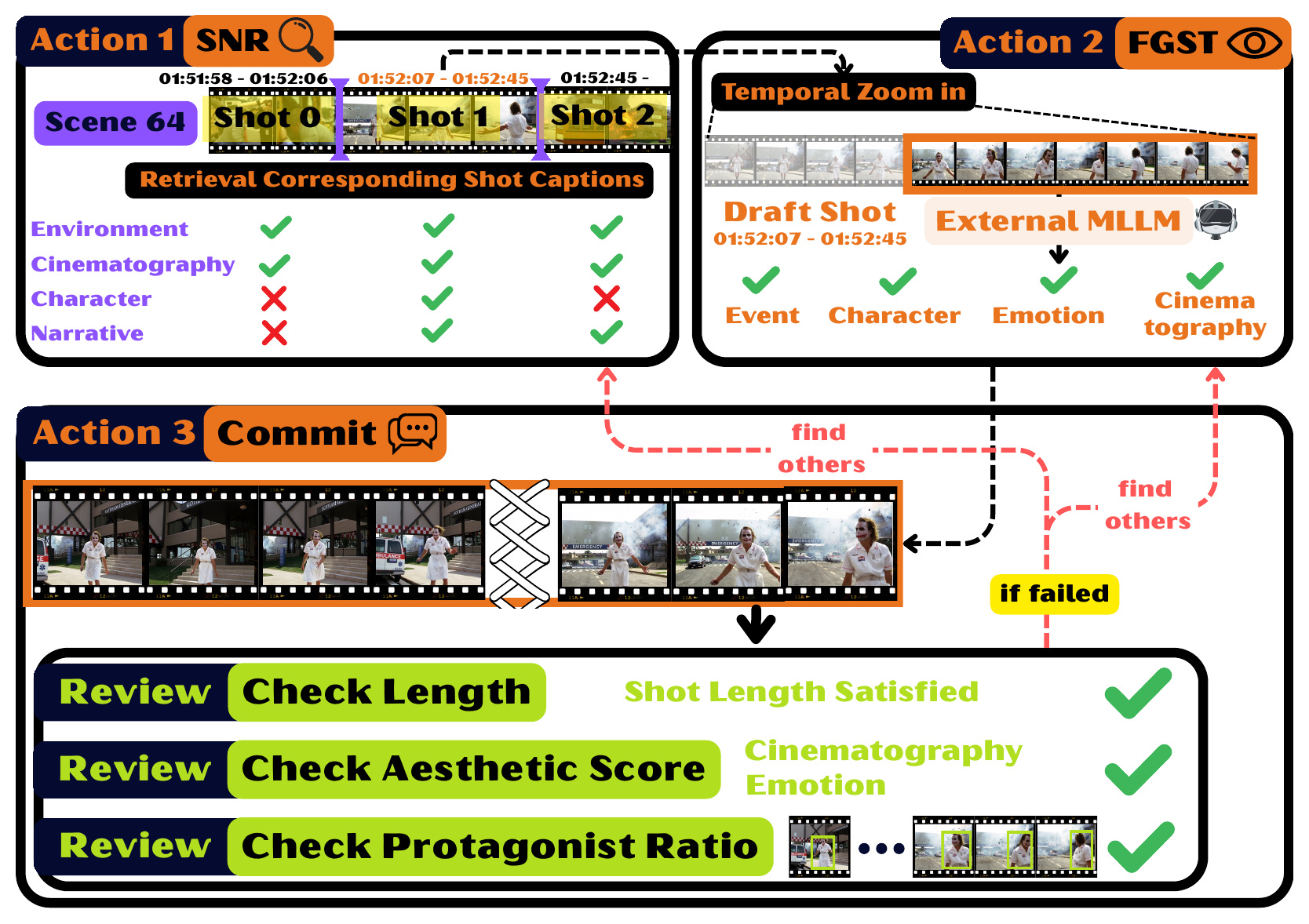}
    % \vspace{-2em}
    \caption{\textit{\textbf{Editor and Reviewer}} are used to perform segment selection and validation. SNR stands for Semantic Neighborhood Retrieval, and FGST stands for Fine-Grained Shot Trimming.}
    \label{fig:editor}
    \vspace{-30pt}
\end{wrapfigure}

As shown in Fig.~\ref{fig:editor}, for each retrieval specification $p_i = (\tau_i, z_{\mathrm{id}}, d_i)$ generated by the Playwriter, the Editor navigates the candidate pool through a hypothesis-and-verification loop. Its goal is to identify a specific clip $c_i = [t^{\mathrm{in}}, t^{\mathrm{out}}]$ such that the duration constraint $|c_i| \approx \tau_i$ is met, while maximizing local utility. The Editor has 3 main actions:

\noindent\textbf{Action 1: Semantic Neighborhood Retrieval.}
This action initializes the local search space $\Omega_i$ by retrieving all shots belonging to the assigned scene $z_{\mathrm{id}}$. To address potential content scarcity or segmentation noise in the visual candidate, we incorporate an Adaptive Expansion mechanism.
If the primary search space $\Omega_i = \{ s \mid s \in z_{\mathrm{id}} \}$ fails to yield a high-confidence candidate, the Editor expands the scope to the semantic neighborhood:
\begin{equation}
    \Omega'_i = \Omega_i \cup \{ s \mid s \in \mathrm{Neighbor}(z_{\mathrm{id}}, \Delta) \}.
\end{equation}
This fallback strategy prevents retrieval dead-ends by aggregating shots from adjacent structural units, ensuring the agent maintains a sufficient material pool for optimization.

\noindent\textbf{Action 2: Fine-Grained Shot Trimming.}
To maximize the objective terms $\mathcal{Q}_{\mathrm{vis}}$ and $\mathcal{Q}_{\mathrm{sync}}$, the Editor employs a VLM-driven analysis tool to perform dense temporal grounding within the candidate shots.
For a candidate shot $s \in \Omega_i$, the agent seeks a sub-segment $c_i \subset s$ that maximizes a weighted local score:
\begin{equation}
    c_i^* = \mathop{\arg\max}_{c \subset s, |c|=\tau_i} \left( \alpha \cdot S_{\mathrm{aes}}(c) + \beta \cdot R_{\mathrm{prot}}(c \mid \mathcal{H}) \right).
\end{equation}
Here, $S_{\mathrm{aes}}$ represents the aesthetic score (contributing to $\mathcal{Q}_{\mathrm{vis}}$), and $R_{\mathrm{prot}}$ denotes the \textit{Protagonist Presence Ratio} (contributing to $\mathcal{Q}_{\mathrm{cond}}$), where $\alpha$ and $\beta$ are the respective balancing weights. The presence ratio is computed by cross-referencing frame content with the character identity set $\mathcal{H}$ established in Sec.~\ref{sec:Video parsing}. If the current segment yields a suboptimal score, the agent heuristically shifts the temporal window based on VLM feedback until a high-fidelity clip is secured.

\noindent\textbf{Action 3: Commit.}
The Editor submits the trimmed candidate $c_i$ to the Reviewer~(Sec.~\ref{sec:Reviewer}). Upon receiving an approval signal, the clip is rendered and committed to the final timeline $\mathcal{E}^*$. Otherwise, the Editor triggers a backtracking mechanism to explore alternative intervals within $\Omega_i$.

\subsection{Reviewer: Multi-Criteria Validity Gate}
\label{sec:Reviewer}

To ensure the final timeline $\mathcal{E}^*$ adheres to both narrative intent and structural constraints, we introduce the Reviewer to operate as a discriminatory gate. As shown in Fig.~\ref{fig:editor}, this module audits every candidate clip $c_i$ proposed by the Editor through a rigorous rejection sampling mechanism. The reviewer checks the consistency of the edited video from the following aspects:

\noindent\textbf{Semantic Identity Verification.}
To enforce narrative consistency ($\mathcal{Q}_{\mathrm{cond}}$), the Reviewer validates that the visual subject strictly aligns with the target identity defined in $\mathcal{H}$. By computing a \emph{Protagonist Presence Ratio} via hierarchical MLLM~\cite{Qwen3-VL} sampling, we filter out false positives where the character is merely a background extra, occluded, or unrecognizable. This ensures that the protagonist remains the primary visual focus throughout the sequence, distinguishing the main characters from crowd elements.

\noindent\textbf{Temporal and Structural Integrity.}
To maintain the topological validity of the timeline, we enforce hard constraints on sequencing. The Reviewer verifies \textit{Non-Overlap} ($\cap \mathcal{E}_{prev} = \emptyset$) to prevent content duplication and checks \textit{Duration Fidelity} to ensure the visual cut points align precisely with the rhythmic grid of the music track $\mathcal{M}$. Any violation of these constraints triggers an immediate rejection to preserve the global structure.

\noindent\textbf{Perceptual Quality Assurance.}
To maximize aesthetic appeal ($\mathcal{Q}_{\mathrm{vis}}$), the module audits low-level visual saliency. It rejects shots exhibiting significant quality degradation, ensuring that every committed segment meets broadcast-level viewing standards. Upon detection of any violation, the Reviewer returns structured feedback, prompting the Editor to backtrack and explore alternative intervals within the semantic neighborhood.

If the candidate clip does not meet the requirements, the Reviewer will notify the editor to select the relevant scene around the current one. By reviewing each candidate clip in the optimized timeline, we obtain the final edited video.
\section{Experiment}

\subsection{Evaluation and Implementation}

\noindent\textbf{Benchmark.}
To rigorously evaluate our framework, we establish a diverse benchmark specifically designed for agentic video editing tasks. Our dataset comprises 10 distinct source pairs, collected from 5 feature-length films and 5 long-duration VLOGs, with raw footage lengths ranging from 1 to 3 hours. This collection accumulates to approximately 24 hours of total footage, ensuring a robust assessment across both professionally cinematographed content and unscripted, naturalistic recordings. The corresponding auditory inputs consist of 10 segmented music tracks spanning a wide spectrum of genres, including Pop, Jazz, OST, Rock, and R\&B with target edit durations varying from 20 seconds to one minute.
    
To test the system's semantic adaptability, we formulate two distinct instruction categories: \textit{(1)Character-Centric Instructions}, which constrain the edit to focus exclusively on a single protagonist, thereby challenging the agent's ability to maintain identity consistency; and \textit{(2)Narrative-Centric Instructions}, which demand the inclusion of multiple characters or complex interactions to convey a cohesive visual story. In total, this benchmark yields 20 unique evaluation cases (10 pairs $\times$ 2 instruction types), covering a broad range of visual styles and narrative requirements.

\noindent\textbf{Metrics.}
We evaluate our framework via automated quantitative analysis and a subjective user study. In the automated regime, \textit{Visual Quality} and \textit{Instruction Follow} are scored by GPT-5.2~\cite{gpt52} based on aesthetic integrity and semantic alignment, respectively. Conversely, given the high temporal precision required for audio-visual alignment, which remains challenging for MLLMs, \textit{AV Harmony} is quantified via detecting the minimum temporal offset $\Delta t$ between audio onsets~(downbeats, pitch) and video scenes, strictly rewarding alignments within a perceptual threshold (e.g., $\Delta t \le 0.1s$). The user study mirrors these three metrics to capture human perceptual preference and exclusively evaluates a fourth dimension, \textit{Human-Likeness}, which benchmarks the naturalness of the model's editing pacing and logic against professional human editors.

% \subsection{Baselines and Implementation Details}
\noindent\textbf{Baselines.}
We benchmark our framework against three representative methods covering different editing paradigms. 
{NarratoAI}~\cite{narratoai} serves as a baseline for subtitle-driven editing; it is a mainstream open-source framework that processes full-video subtitles to generate clips based on textual instructions. Note that NarratoAI is inapplicable to VLOG scenarios due to the scarcity of speech and subtitles in such footage.
{UVCOM} \cite{uvcom} and {Time-R1} \cite{timer1} represent state-of-the-art approaches in moment retrieval and temporal grounding, respectively. Since both models typically handle fixed-length short videos, we adapt them for long-form footage by first segmenting the source video, then selecting the top-5 clips with the highest confidence. Finally, we trim the selected segments to match the target duration, discarding excess frames.

\noindent\textbf{Implementation Details.}
For the core agentic framework, we employ MiniMax-M2.1~\cite{MiniMaxM21} to power the Editor and Reviewer agents, while Gemini3-Pro~\cite{google2025gemini3} serves as the Playwriter. 
In the preprocessing stage, we utilize PySceneDetect~\cite{pyscenedetect} for shot boundary detection and Whisper-v3-turbo \cite{whisper} for Automatic Speech Recognition (ASR) to extract subtitles. 
For multimodal understanding, Qwen3-VL-30B-A3B~\cite{Qwen3-VL} and Qwen3-Omni-30B-A3B~\cite{Qwen3-Omni} are deployed for visual and music captioning, respectively. 
To optimize computational efficiency during inference, video footage is downsampled to a short-side resolution of 360p at a frame rate of 2 FPS.

\begin{figure*}[!t]
\centering
  \includegraphics[width=\textwidth]{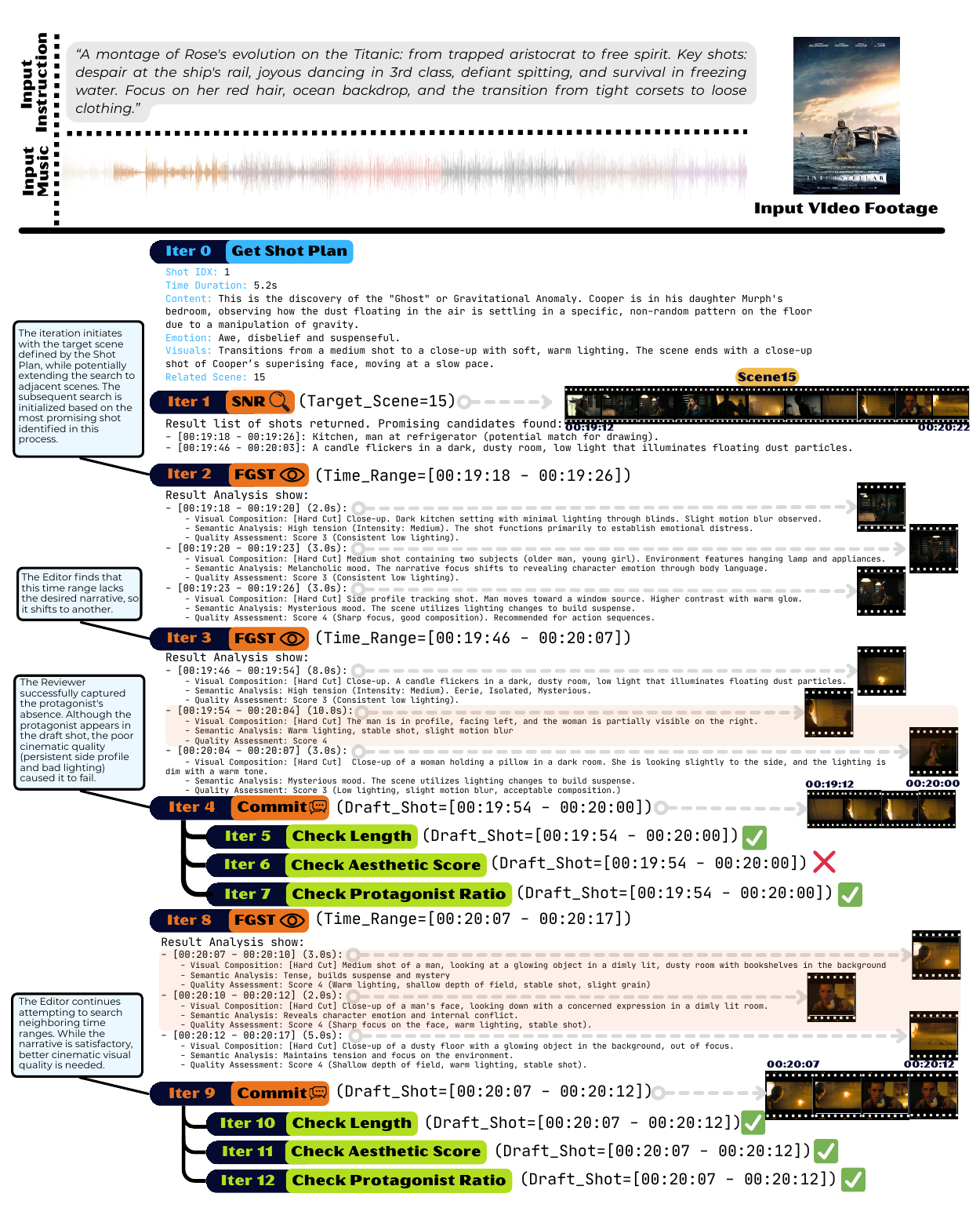}
  \vspace{-3em}
    \caption{\textbf{\textit{A sample execution of a single-shot cutting}}, utilizing footage from movie ``Interstellar'' and the music ``Moon.'' Actions performed by the Playwright, Editor, and Reviewer are color-coded in \textcolor[RGB]{96,180,249}{blue}, \textcolor[RGB]{219,122,55}{yellow}, and \textcolor[RGB]{187,221,77}{green}, respectively. The \colorbox[RGB]{251,241,233}{orange background} traces the execution path leading to the final clip selection.}
  \label{fig:workflow}
  \vspace{-2em}
\end{figure*}

\subsection{Main Results}
\begin{figure*}[!t]
\centering
  \includegraphics[height=0.97\textwidth]{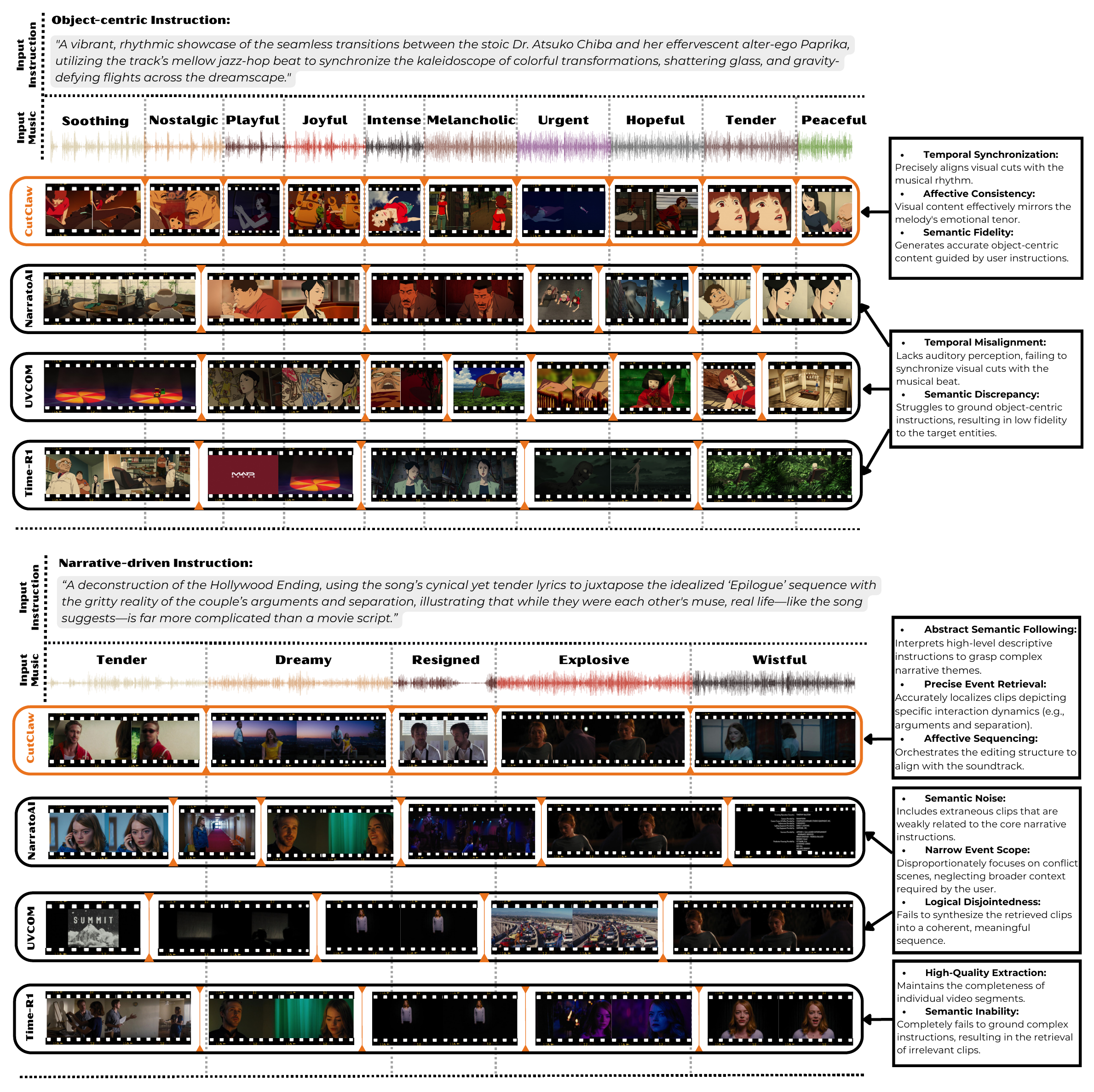}
  \vspace{-1em}
    \caption{\textbf{\textit{Qualitative comparison between CutClaw and baseline methods.}} The two cases utilize full-length footage from the films ``Paprika'' and ``La La Land'', paired with the musical tracks ``Luv(sic) Pt.2'' and ``Norman F**king Rockwell'', respectively. Shot boundary detection is performed using PySceneDetect~\cite{pyscenedetect}.}
  \label{fig:cmoparison}
  \vspace{-2em}
\end{figure*}

As presented in Tab.~\ref{tab:main}, CutClaw achieves superior performance across all quantitative metrics. Validating our coarse-to-fine editing strategy, our method consistently surpasses the strongest baselines in Visual Quality. Furthermore, CutClaw leads the Instruction Follow metric, particularly excelling in object-oriented instruction, which demonstrates its precise visual content localization capabilities. Finally, its dominant performance in AV Harmony confirms that the resulting cuts are rigorously and rhythmically aligned with the input music. Fig.~\ref{fig:cmoparison} shows the qualitative comparison against baseline methods. When considering the overall edited quality, baseline methods exhibit rigid segment selection, completely failing to align with the musical structure. While NarratoAI loosely follows user instructions at the cost of severe visual degradation, UVCOM and Time-R1 maintain visual quality but lack logical narrative connections across shots. Additional video results are available in the supplementary material. To further illustrate our method, we provide an execution sample in Fig.~\ref{fig:workflow}, detailing the collaborative workflow among the Playwriter, Editor, and Reviewer agents.
\vspace{-1em}
\par\begin{table*}[b]
    \centering
    \begin{minipage}[t]{0.49\textwidth}
        \centering
        \centering
\captionof{table}{\textit{\textbf{Comparision}}. We report the performance scores across three metrics: Visual Quality, Instruction Follow (Obj/Nar), and AV Harmony.}
\vspace{-1em}
\label{tab:main}
\resizebox{\columnwidth}{!}{
\setlength{\tabcolsep}{4pt} 
\begin{tabular}{l ccc ccc ccc}
\toprule
\multirow{2}{*}{\textbf{Method}} & \multicolumn{3}{c}{\textbf{Visual Quality}} & \multicolumn{3}{c}{\textbf{Instruction Follow}} & \multicolumn{3}{c}{\textbf{AV Harmony}} \\
\cmidrule(lr){2-4} \cmidrule(lr){5-7} \cmidrule(lr){8-10}
 & Film & Vlog & Avg. & Obj & Nar & Avg. & Film & Vlog & Avg. \\
\midrule
NarratoAI & 75.7 & - & 75.7 & 56.0 & 72.0 & 64.0 & 84.9 & - & 84.9 \\
UVCOM & 71.2 & 73.6 & 72.4 & 60.8 & 64.5 & 62.6 & 78.9 & 79.7 & 79.3 \\
Time-R1 & 73.3 & 72.6 & 72.9 & 51.9 & 71.0 & 61.5 & 77.0 & 75.8 & 76.4 \\
\midrule
\textbf{CutClaw} & \textbf{79.2} & \textbf{76.0} & \textbf{77.6} & \textbf{66.6} & \textbf{73.4} & \textbf{70.0} & \textbf{85.7} & \textbf{87.3} & \textbf{86.5} \\
\bottomrule
\end{tabular}
}
\vspace{-1em}
    \end{minipage}\hfill
    \begin{minipage}[t]{0.49\textwidth}
        \centering
        \centering
\captionof{table}{\textbf{\textit{Ablation Study}}. We report the performance impact of ablating the Editor, Reviewer, or Audio Context across three metrics.}
\vspace{-1em}
\label{tab:abl}
\resizebox{\columnwidth}{!}{
\setlength{\tabcolsep}{3pt} 
\begin{tabular}{l ccc ccc ccc}
\toprule
\multirow{2}{*}{\textbf{Method}} & \multicolumn{3}{c}{\textbf{Visual Quality}} & \multicolumn{3}{c}{\textbf{Instruction Follow}} & \multicolumn{3}{c}{\textbf{AV Harmony}} \\
\cmidrule(lr){2-4} \cmidrule(lr){5-7} \cmidrule(lr){8-10}
 & Film & Vlog & Avg. & Obj & Nar & Avg. & Film & Vlog & Avg. \\
 
\midrule
w/o Audio & 77.3 & 73.8 & 75.5 & 63.4 & \textbf{74.3} & 68.9 & 78.0 & 76.5 & 77.2 \\
w/o Editor & \underline{78.6} & \underline{75.4} & \underline{77.0} & 59.7 & 71.5 & 65.6 & 84.8 & 86.0 & 85.4 \\
w/o Reviewer & 78.1 & 74.0 & 76.0 & \textbf{66.6} & 72.9 & \underline{69.8} & \underline{85.1} & \textbf{89.4} & \textbf{87.2} \\
\midrule
\textbf{CutClaw} & \textbf{79.2} & \textbf{76.0} & \textbf{77.6} & \textbf{66.6} & \underline{73.4} & \textbf{70.0} & \textbf{85.7} & \underline{87.3} & \underline{86.5} \\
\bottomrule
\end{tabular}
}
\vspace{-2em}

    \end{minipage}
\end{table*}\par

\subsubsection{Ablation Study}
To validate the effectiveness of individual components within CutClaw, we conducted an ablation study by systematically removing the Editor, Reviewer, and Audio Context, with results detailed in Tab.~\ref{tab:abl}. We first observe that replacing the Audio's beat-aware analysis with fixed-length segmentation causes AV Harmony to drop significantly from 86.5 to 77.2, confirming its necessity for rhythmic alignment. Similarly, removing the Reviewer leads to a decline in Visual Quality from 77.6 to 76.0, as the system loses the feedback loop required to refine low-quality candidates and transition mismatches. Finally, substituting the Editor with a random clip selector degrades performance across both Visual Quality and Instruction Following, reducing the average score from 70.0 to 65.6. This demonstrates that the Editor's hierarchical structuring is foundational for preserving narrative coherence and semantic accuracy.

\begin{table*}[t]
\centering
\caption{\textbf{\textit{User Study Results}}. We report the percentage of user votes across four metrics: Visual Quality, Instruction Following, Audio-Visual Harmony, and Human-Likeness. For each metric, we break down performance by instruction type (Narrative/Object) and video type (Film/Vlog). Avg. denotes the average score. Our method outperforms baselines across all categories. * Note that NarratoAI cannot deal with the VLOG as it does not have dense subtitles.}
\vspace{-1em}
\label{tab:user_study_full}
\resizebox{\textwidth}{!}{
\begin{tabular}{l ccccc ccccc ccccc ccccc}
\toprule
\multirow{2}{*}{\textbf{Method}} & \multicolumn{5}{c}{\textbf{Visual Quality}} & \multicolumn{5}{c}{\textbf{Instruction Follow}} & \multicolumn{5}{c}{\textbf{Audio-Visual Harmony}} & \multicolumn{5}{c}{\textbf{Human-Like}} \\
\cmidrule(lr){2-6} \cmidrule(lr){7-11} \cmidrule(lr){12-16} \cmidrule(lr){17-21}
 & Nar & Obj & Film & Vlog & Avg & Nar & Obj & Film & Vlog & Avg & Nar & Obj & Film & Vlog & Avg & Nar & Obj & Film & Vlog & Avg \\
\midrule
NarratoAI* & 11.6\% & 11.2\% & 27.3\% & - & 11.4\% & 14.8\% & 10.8\% & 28.7\% & - & 12.8\% & 11.2\% & 12.4\% & 25.3\% & - & 11.8\% & 12.0\% & 8.4\% & 23.3\% & - & 10.2\% \\
UVCOM & 18.0\% & 16.8\% & 7.3\% & 23.6\% & 17.4\% & 18.8\% & 13.2\% & 7.3\% & 22.0\% & 16.0\% & 17.2\% & 13.2\% & 6.0\% & 22.8\% & 15.2\% & 18.8\% & 15.6\% & 9.3\% & 23.6\% & 17.2\% \\
Time-R1 & 25.2\% & 17.6\% & 18.0\% & 22.4\% & 21.4\% & 22.4\% & 19.6\% & 16.7\% & 24.0\% & 21.0\% & 23.2\% & 16.8\% & 17.3\% & 18.4\% & 20.0\% & 24.8\% & 22.8\% & 16.7\% & 25.6\% & 23.8\% \\
\midrule
\textbf{CutClaw} & \textbf{45.2\%} & \textbf{54.4\%} & \textbf{47.3\%} & \textbf{54.0\%} & \textbf{49.8\%} & \textbf{44.0\%} & \textbf{56.4\%} & \textbf{47.3\%} & \textbf{53.6\%} & \textbf{50.2\%} & \textbf{48.4\%} & \textbf{57.6\%} & \textbf{51.3\%} & \textbf{57.6\%} & \textbf{53.0\%} & \textbf{44.4\%} & \textbf{53.2\%} & \textbf{50.7\%} & \textbf{50.4\%} & \textbf{48.8\%} \\
\bottomrule
\end{tabular}
}
\vspace{-2em}
\end{table*}

\subsubsection{User Study}
To complement the objective metrics, we conducted a user preference study to assess the subjective quality of the generated videos. We recruit 25 participants to evaluate the results. The questionnaire consists of 80 evaluation items, asking participants to vote for the best method across four dimensions: Visual Quality, Instruction Follow, Audio-Visual Harmony, and Human-Likeness. In total, we collected 2,000 user opinions, providing a statistically robust basis for our analysis. As illustrated in Tab.~\ref{tab:user_study_full}, CutClaw outperforms all baselines by a significant margin across all categories. Specifically, our method receive 49.8\% of the votes for Visual Quality and 53.0\% for Audio-Visual Harmony, which is more than double the votes received by the second-best method, Time-R1 (21.4\% and 20.0\%, respectively). Notably, in the Human-Like metric, CutClaw secured 48.8\% of user preference, highlighting its ability to mimic professional human editing logic better than existing automated solutions. These results align consistently with our quantitative findings, confirming the superiority of our approach in real-world viewer evaluations.

\subsection{Limitation}
Our framework still faces limitations. First, while we ensure strong narrative flow, the system lacks advanced visual hooks, such as generated visual effects or specific monologue highlights that are crucial for engaging content. Future iterations could integrate generative video models to synthesize these expressive elements. Second, the multi-stage pipeline processing extensive raw footage results in high inference latency. Optimizing the pipeline for speed or employing coarse-to-fine processing strategies to enable real-time feedback remains a key direction for future research.

\section{Conclusion}

We presented CutClaw, an autonomous multi-agent framework designed to automate the complex task of professional video editing from hours-long raw footage. By addressing the critical challenges of processing long contexts and achieving precise audio-visual consistency, our approach bridges the gap between simple clip assembly and instruction-aligned, music-driven storytelling. The core innovation of our framework lies in its hierarchical decomposition strategy, which transforms continuous high-dimensional data into structured semantic units. This structure allows our specialized agents to collaborate effectively: the Playwriter anchors the narrative to musical structure, the Editor performs fine-grained visual grounding, and the Reviewer enforces rigorous aesthetic and continuity constraints. 
Our extensive experimental results demonstrate that CutClaw significantly outperforms state-of-the-art baselines across key metrics, including Visual Quality, Instruction Following, and AV Harmony. 
% Furthermore, subjective user studies confirm that our system generates videos with a level of rhythmic synchronization and logical pacing that users perceive as highly human-like. SoniCut represents a robust step forward in intelligent content creation, offering a scalable solution for transforming massive volumes of visual data into compelling, professional-grade videos.

\section*{Acknowledgements}
This work was financially supported in part by the National Natural Science Foundation of China (Project No. 62506064) and Guangdong Provincial Regional Joint Fund (Project No. 2024A1515110052). The computational resources are supported by SongShan Lake HPC Center (SSL-HPC) in Great Bay University.

\bibliographystyle{splncs04}
\bibliography{main}

\end{document}